\documentclass[runningheads]{llncs}
\usepackage[T1]{fontenc}
\usepackage{graphicx}

\usepackage{booktabs}
\usepackage{multirow}
\usepackage{makecell}
\usepackage{amsmath}
\usepackage{cite}

\begin{document}
\title{Mix and Match: Context Pairing for Scalable Topic-Controlled Educational Summarisation}
\titlerunning{Context Pairing for Scalable Topic-Controlled Educational Summarisation}

\author{Nathikan Yodthapa\inst{1} \and Thanapong Intharah\inst{1}
\and Sahan Bulathwela\inst{2}\thanks{Corresponding author}
}
\authorrunning{N. Yodthapa et al.}
\institute{Visual Intelligence Laboratory, Department of Statistics, Faculty of Science, Khon Kaen University, Khon Kaen 40002, Thailand \\
\email{nathikan@kkumail.com, thanin@kku.ac.th}
\and
Centre for Artificial Intelligence, Department of Computer Science, University College London, London WC1E 6BT, The United Kingdom \\
\email{m.bulathwela@ucl.ac.uk}}
\maketitle
\begin{abstract}
Topic-controlled summarisation enables users to generate-summaries focused on specific aspects of source documents. This paper investigates a data augmentation strategy for training small language models (sLMs) to perform topic-controlled summarisation.  We propose a pairwise data augmentation method that combines contexts from different documents to create contrastive training examples, enabling models to learn the relationship between topics and summaries more effectively. Using the SciTLDR dataset enriched with Wikipedia-derived topics, we systematically evaluate how augmentation scale affects model performance. Results show consistent improvements in win rate and semantic alignment as the augmentation scale increases, while the amount of real training data remains fixed. Consequently, a T5-base model trained with our augmentation approach achieves competitive performance relative to larger models, despite using significantly fewer parameters and substantially fewer real training examples.

\end{abstract}

\section{Introduction}

The growing volume of academic literature creates significant challenges for learners, particularly in science education, where students must efficiently access information aligned with specific learning objectives. For instance, a student studying neural networks may only need to understand how backpropagation works from a paper, rather than reading the entire work. Generic summaries often fail to meet these needs, as they provide broad overviews rather than being focused on targeted topics. Topic-controlled summarisation addresses this limitation by enabling summaries to be generated with explicit topical constraints, 
reducing cognitive load and supporting more effective learning.

Although large language models (LLMs) have demonstrated strong performance in summarisation tasks, their high computational and data requirements limit their applicability in many educational settings. Smaller language models (sLMs) provide a more practical alternative; however, their performance is particularly sensitive to the quality and availability of labelled training data. In educational domains, high-quality labelled data is often scarce. As a result, improving data efficiency by maximising the utility of limited high-quality data becomes essential for enabling effective topic-controlled summarisation with sLMs.

In this work, we propose a pairwise data augmentation strategy that improves data efficiency for topic-controlled summarisation by constructing contrastive training examples while keeping the amount of real training data fixed. We conduct a controlled study on the augmentation scale, demonstrating consistent improvements in semantic alignment and relative topical alignment as augmentation increases, with stable performance at higher scales. Under limited real-data conditions, we further show that a T5-base model trained with our approach achieves performance comparable to larger topic-controllable summarisation models, despite using substantially fewer parameters and training examples, highlighting its suitability for resource-constrained educational applications.

\subsubsection{Research Questions}
The research questions are formulated as follows:

\textbf{RQ1:} Can pairwise data augmentation improve topic-controlled 
summarisation under limited data conditions? 

\textbf{RQ2:} How does 
augmentation scale affect topic alignment as measured by win rate? 

\textbf{RQ3:} Can a small language model trained with this strategy 
achieve performance comparable to larger models?

\section{Related Work}

Recent advances in transformer-based architectures \cite{vaswani2017} have substantially improved abstractive text summarisation. Building on this foundation, pretrained models such as T5 \cite{raffel2020}, BART \cite{lewis2020}, and PEGASUS \cite{zhang2020pegasus} achieve strong performance on standard benchmarks by generating fluent, human-like summaries. These models have further enabled controllable summarisation, where generation is guided toward specific topics or attributes. For example, controllable summarisation methods guide generation toward specific topics using explicit control mechanisms. Early approaches explore entity-planning strategies \cite{liu2021}, followed by control-code-based methods such as CTRLsum \cite{he2022}, and more recent work on query-based control \cite{chen2023}. These methods are effective but are commonly implemented on large pretrained models and require substantial labelled data, which can limit their practicality in resource-constrained educational settings.

In educational contexts, topic-controlled generation is particularly valuable for aligning generated content with learners’ objectives. Li et al. propose topic-controlled educational question generation using scalable topic annotation via Wikification, demonstrating the potential of topic guidance for personalised learning. While this line of work targets question generation rather than summarisation, it highlights the importance of topic control in educational applications. In parallel, data augmentation techniques have been explored to address data 
scarcity in summarisation. For instance, \cite{loem2022} propose 
ExtraPhrase, which constructs pseudo training pairs via extractive summarisation followed by paraphrasing, demonstrates effectiveness in low-resource settings. 
However, such approaches focus on improving surface-level diversity rather than reinforcing specific topic--summary associations. In the aspect-based 
summarisation setting, \cite{yang2023oasum} addresses data scarcity by constructing a large-scale dataset from Wikipedia to enable aspect-focused 
generation. While effective, this approach relies on large-scale data collection rather than augmenting limited labelled data. In contrast, our work introduces 
a pairwise augmentation strategy adapted from \cite{li2025} that explicitly targets topic--summary alignment by exposing the model to contrastive contexts, 
enabling effective training of small language models without requiring additional labelled data or large-scale corpora.

\section{Methodology}

\subsection{Models}
We use T5-base as the foundation model due to its compact size and strong empirical performance in summarisation. T5 is a transformer-based encoder–decoder model with approximately 220 million parameters, making it suitable for conditional generation under resource constraints. Input sequences are truncated to 512 tokens, and generated summaries are limited to 128 tokens. For RQ 3 experiments, we use CTRLSum \cite{he2022}, a strong topic-controlled summarisation model that is twice as large as T5-base (406M params). 

\subsection{Dataset Construction}
\textbf{Source Dataset.} We base our experiments on the SciTLDR-A corpus, which contains scientific paper abstracts paired with expert- or author-written TLDR-style summaries \cite{cachola2020}. The dataset is split into training (1,992 abstracts), validation (618), and test (619) sets. While the validation and test splits may include multiple reference summaries per abstract, each training instance contains a single abstract–summary pair. Importantly, SciTLDR-A does not provide explicit topic annotations, which are required for topic-controlled summarisation.

\textbf{Topic Annotation.}
To enable topic-controlled training, we augment the training split with topic labels using Wikification \cite{brank2017}, following prior work on topic-guided generation \cite{li2025}. Wikification maps text to relevant Wikipedia concepts and assigns confidence scores based on graph-based ranking. For each abstract--summary pair $(A_i, s_i)$, where $A_i$ denotes the abstract and $s_i$ the corresponding summary, we select a salient topic $t_i$, resulting in topic-annotated training instances of the form $(A_i, t_i, s_i)$. During evaluation, $t'_i$ denotes an alternative topic drawn from the same document, used as a contrastive reference in win rate computation.

\textbf{New Datasets with Novel Augmentation:} For the experiments, we create three versions of datasets from the source SciTLDR training data using the exact method used in \cite{li2025}, but instead of questions as labels, we have summaries in this work. 
From training data that has the following composition
\begin{align}
  D \in \{\dots, (A_i, s_i), (A_j, s_j), \dots\}  \nonumber
\end{align}

We create three new training datasets using topic annotation with Wikification \cite{brank2017}. The dataset families are shown in Table~\ref{tab_data}.

\begin{table}[h]
\caption{The resultant structure of training examples in different training datasets $D_{\ \cdot\ }$ under different augmentation settings. $+$ indicates the concatenation of two abstracts.}
\label{tab_data}
\centering
\footnotesize
\resizebox{\linewidth}{!}{
\begin{tabular}{l c}
\hline
Dataset Name & Dataset Composition \\
\hline
Baseline & $D_{\text{Baseline}} \in \{\dots, (A_i, t_i, s_i), (A_j, t_j, s_j), \dots\}$\\
TCS $\cdot$ X & $D_{\text{TCS1X}} \in \{\dots, (A_i + A_j, t_i, s_i), (A_i + A_j, t_j, s_j), \dots\}$\\
TCS $\cdot$ XX & $D_{\text{TCS1XX}} \in \{\dots, (A_i + A_j, t_i, s_i), (A_i + A_j, t_j, s_j),
(A_j + A_i, t_i, s_i), (A_j + A_i, t_j, s_j), \dots\}$\\
\hline
\end{tabular}
}
\end{table}

\subsection{Evaluation Metrics}
We evaluate models using two automatic metrics for summary quality and topic alignment, respectively:

\textbf{BERTScore}, an indicator of summary quality, measures semantic similarity between the generated and reference summaries using contextualised embeddings, allowing for a flexible comparison beyond surface-level token overlap.

\textbf{Win Rate} evaluates topic alignment between the topic $t_i$ and summary $s_i$ using cosine similarity computed over sentence embeddings. Specifically, we encode the generated summary and topic representations using a pretrained sentence embedding model, and compare the cosine similarity between the generated summary and its intended reference salient topic $\cos(\text{emb}(s_i), \text{emb}(t_i))$ against similarity with the next alternative topic in the same summary $\cos(\text{emb}(s_i), \text{emb}(t'_i))$. A win is recorded when the similarity to the intended topic $t_i$ is higher than that to the alternating topic $t'_i$. The win rate is defined as the proportion of such wins across all evaluated instances. The alternative topic $t'_i$ is derived from the structure of the SciTLDR test set, where each record contains multiple expert-written summaries with corresponding topics. We expand each record into multiple instances, each using one summary-topic pair $(s_i,t_i )$ as the target, while the remaining topics serve as alternative topics $t'_i$. This design reflects realistic evaluation conditions, as the alternative topics are semantically close to the intended topic, making the distinction more challenging and better aligned with real-world data distribution. Although cosine similarity is used as the underlying measure, win rate provides a relative comparison that focuses on topic control rather than direct similarity to a single gold label summary.

\subsection{Experimental Setup}
Table~\ref{tab:exp_setup} summarises the experimental configuration used across all models. To ensure fair comparison under resource-constrained settings, all experiments share the same training and optimisation setup.

The test data split of the SciTLDR dataset contains multiple summaries created by several experts (author, domain expert, etc.) for each unique example. Due to this, we do not carry out abstract pairing in the test set and use different summaries that have different salient topics as topic-controlled examples. This allows us to do a truly realistic evaluation that aligns with the real-world data distribution.

\begin{table}[h]
\centering
\caption{Experimental setup and training configuration for T5 Models}
\begin{tabular}{l l}
\hline
Component & Setting \\
\hline
Model & T5-base encoder--decoder \\
Input format & \textit{Summarize: topic = $t_i$, context = $A_i + A_j$} \\
Max input length & 512 tokens \\
Max output length & 128 tokens \\
Optimiser & AdamW (weight decay 0.05) \\
Learning rate & $5 \times 10^{-6}$ with linear warm-up (10\%) \\
Batch size & 8 \\
Training epochs & Up to 10 \\
Loss function & Sequence-to-sequence cross-entropy \\
Regularisation & Dropout (0.3 hidden, 0.1 attention) \\
Gradient clipping & 1.0 \\
Early stopping & Validation loss (patience 1, $\delta = 10^{-4}$) \\
\hline
\end{tabular}
\label{tab:exp_setup}
\end{table}

\section{Results}

\subsection{Impact of Data Augmentation on Model Performance (RQ1)}
Table~\ref{tab_base} reports the performance of the baseline model and its augmented variants under small-scale topic-controlled sampling (TCS). Using 1× augmentation (TCS1X) does not lead to clear improvements over the baseline in either BERTScore or win rate. Although TCS1XX introduces reversed-context samples to increase diversity, the resulting gains remain marginal, with win rate matching the baseline and only slight stabilisation in BERTScore.

When the augmentation scale is increased to 2× (TCS2X), a modest but consistent improvement in win rate is observed, while BERTScore remains comparable across all settings. This indicates that limited augmentation alone is insufficient to substantially improve topic alignment, whereas moderate increases in augmented data begin to yield measurable benefits without degrading summary quality.

\begin{table}[h]
\centering
\normalsize
\caption{Results of summarisation models under different training data settings. The best and second-best results are indicated in \textbf{bold} and \emph{italic} respectively.}
\label{tab_base}
\begin{tabular}{l c c}
\hline
\textbf{Model} 
& \textbf{BERTScore} 
& \textbf{Win rate (\%)} \\
\hline
\textbf{Baseline} & \textbf{0.8795} & $\mathit{36.56}$ \\
\textbf{TCS1X} & 0.8779 $\pm$ 0.0004 & 35.69 $\pm$ 1.0988 \\
\textbf{TCS1XX} & $\mathit{0.8783\pm 0.0007}$ & $\mathit{36.56 \pm 1.0825}$ \\
\hline
\textbf{TCS2X} & 0.8780 $\pm$ 0.0003 & \textbf{36.80 $\pm$ 0.5098} \\
\hline
\end{tabular}
\end{table}


\subsection{Effects of Augmentation Scale on Pairwise Topic-Controlled summarisation (RQ2)}
Table~\ref{tab:bert_tcs_comparison} presents model performance across a wider range of TCS augmentation scales. BERTScore remains relatively stable across all configurations, suggesting that increasing the amount of augmented data does not negatively affect overall summary quality. In contrast, the win rate shows a clearer upward trend as the augmentation scale increases, indicating progressively better topic alignment.

Improvements in win rate are already observable at small augmentation scales (2x to 3x) and continue to increase at larger scales, peaking at 15x–30x. Notably, TCS1X performs below the baseline, and win rate improvements only become more apparent from TCS3X onward, suggesting that a certain level of contrastive exposure may be needed for the augmentation to take effect. The decreasing standard deviation at larger scales may also suggest improved training stability.

Despite relying on a substantially smaller model and significantly fewer real training instances, TCS30X achieves a win rate comparable to CTRLsum, which uses a larger backbone and a much larger labelled dataset. This comparison suggests that pairwise topic-controlled augmentation can effectively improve topic alignment under limited data and model capacity constraints.

\begin{table}[ht]
\centering
\caption{Performance increase when models are trained with new datasets created by the proposed data augmentation method in different multipliers of the original training dataset size, ultimately leading to beating CTRLSum with no additional examples.}
\label{tab:bert_tcs_comparison}
\begin{tabular}{lcccc}
\hline
\textbf{Model} & \textbf{Params} & \textbf{\# Unique Training Examples} & \textbf{BERTScore} & \textbf{Win rate (\%)} \\
\hline
Baseline & 220M & 1,992 & 0.8795 & 36.56 \\
TCS1X & 220M & 1,992 & 0.8779 $\pm$ 0.0004 & 35.69 $\pm$ 1.0988 \\
TCS2X & 220M & 1,992 & 0.8780 $\pm$ 0.0003 & 36.80 $\pm$ 0.5098 \\
TCS3X & 220M & 1,992 & 0.8783 $\pm$ 0.0002 & 38.22 $\pm$ 0.1651 \\
TCS4X & 220M & 1,992 & 0.8783 $\pm$ 0.0001 & 37.51 $\pm$ 0.2423 \\
TCS5X & 220M & 1,992 & 0.8781 $\pm$ 0.0002 & 37.43 $\pm$ 1.0041 \\
TCS10X & 220M & 1,992 & 0.8789 $\pm$ 0.0003 & 38.38 $\pm$ 0.2859 \\
TCS15X & 220M & 1,992 & 0.8789 $\pm$ 0.0003 & 38.86 $\pm$ 0.1651 \\
TCS25X & 220M & 1,992 & 0.8714 $\pm$ 0.0146 & 39.81 $\pm$ 0.2747 \\
TCS30X & 220M & 1,992 & \textbf{0.8799 $\pm$ 0.0001} & \textbf{39.97 $\pm$ 0.5200} \\
CTRLsum & 406M & 50,000 & 0.8545 & \textbf{39.97} \\
\hline
\end{tabular}
\end{table}

\section{Discussion}
Our results indicate that pairwise augmentation improves topic-controlled summarisation primarily by increasing contextual diversity rather than merely enlarging the training set. As shown in Table~\ref{tab_base}, the basic paired setting (TCS1X) does not outperform the baseline, suggesting that naive pairing alone is insufficient. Introducing reversed-context samples (TCS1XX) stabilises performance, while moderate scaling (TCS2X) leads to measurable improvements in win rate, indicating better topic alignment. This further suggests that exposing the model to diverse paired contexts helps it learn more robust topic–summary associations.

The observed pattern across augmentation scales offers further insight into how pairwise augmentation affects model learning. At low scales (1x-2x), the paired contexts may introduce additional noise without providing sufficient contrastive signal, which could explain why the model does not consistently outperform the baseline. As the scale increases (3x-10x), a greater variety in topic--context pairings may help the model develop more robust topic--summary associations. At larger scales (15x-30x), the accumulated contrastive signal appears to stabilise training, as reflected in both higher win rates and reduced variance across runs, though the underlying mechanism warrants further investigation.


Compared to CTRLsum, which uses a larger model and labelled data, our T5-base models achieve similar win rates with far fewer parameters and only 1,992 documents. While CTRLsum benefits from greater capacity and supervision, these results suggest that pairwise topic-controlled augmentation can help small models approach the performance of larger models under data constraints. From an educational perspective, higher win rates indicate that the model’s summaries align more closely with the intended topic, supporting learners’ objectives and potentially reducing instructor workload and learner cognitive overload in Intelligent Tutoring Systems.

Unlike prior work, such as ExtraPhrase \cite{loem2022} and OASum \cite{yang2023oasum}, which emphasise data diversity, our gains occur without changes in BERTScore, suggesting that pairwise augmentation primarily improves topic alignment rather than surface-level diversity. Explicit modelling of contrastive topic–context relationships may thus aid topic-controlled summarisation in low-resource settings.

\section{Conclusion}

This study demonstrates that pairwise data augmentation can effectively train a T5-base model for topic-controlled summarisation under limited real-data conditions. Our results show that augmentation improves performance primarily by increasing contextual diversity rather than merely enlarging the dataset. As the augmentation scale increases, win rate consistently improves, indicating stronger topic alignment, while BERTScore remains stable across settings, suggesting preserved semantic quality.

At larger augmentation scales, gains in win rate continue without sacrificing overall summary quality, highlighting the robustness of the proposed pairwise training strategy. Compared with CTRLsum, our approach achieves comparable topic alignment using a substantially smaller model and only 1,992 real training examples. These findings underscore the data efficiency of pairwise topic-controlled augmentation and suggest its potential as a practical alternative for controllable summarisation in resource-constrained settings. Future work will explore alternative augmentation strategies and evaluate generalisation across different domains and datasets.

\section{Limitation and future work}
Several limitations should be noted. Evaluation relies solely on automatic metrics, and human evaluation has not been conducted. Formal statistical significance testing was not performed, and results should be interpreted 
with caution. Comparisons are limited to CTRLsum; benchmarking against general-purpose LLMs, broader topic annotation methods, and cross-domain generalisation remains as future work.


\begin{credits}
\subsubsection{\ackname} 
This work is co-funded by the European Commission’s projects “Teacher-AI Complementarity (TaiCo)" (Project ID: 101177268), “Humane AI" (Grant No. 820437) and “X5GON" (Grant
No. 761758).

\end{credits}

\bibliographystyle{splncs04}

\end{document}